\documentclass{article}
\usepackage{graphicx}
\usepackage{color, colortbl}
\usepackage{arydshln}
\usepackage{booktabs}
\usepackage{spconf,amsmath,epsfig}
\usepackage{amssymb}
\usepackage{multicol,tabularx,capt-of}
\usepackage{multirow}
\usepackage{hyperref}
\hypersetup{colorlinks=true, linkcolor=black, filecolor=black, urlcolor=black}

\usepackage[autostyle=false, style=english]{csquotes}
\definecolor{LightCyan}{rgb}{0.85,1,1}
\definecolor{Gray}{gray}{0.95}
\MakeOuterQuote{"}
\makeatletter
\newcommand{\ostar}{\mathbin{\mathpalette\make@circled\star}}
\newcommand{\make@circled}[2]{%
  \ooalign{$\m@th#1\smallbigcirc{#1}$\cr\hidewidth$\m@th#1#2$\hidewidth\cr}%
}
\newcommand{\smallbigcirc}[1]{%
  \vcenter{\hbox{\scalebox{0.77778}{$\m@th#1\bigcirc$}}}%
}
\makeatother

\renewcommand\footnotemark{}

\title{IMAGE COMPLETION VIA DUAL-PATH COOPERATIVE FILTERING}

\name{Pourya Shamsolmoali$^{1}$, Masoumeh Zareapoor$^2$, Eric Granger$^3$}
\address{$^1$Shanghai Key Laboratory of Multidimensional Information Processing, East China Normal University, China\\
$^2$School of Automation, Shanghai Jiao Tong University, China\\
$^3$Lab. d’imagerie, de vision et d’intelligence artificielle, Dept. of Systems Eng., ETS, Canada}

\begin{document}

\maketitle

\begin{abstract}
Given the recent advances with image-generating algorithms, deep image completion methods have made significant progress. However, state-of-art methods typically provide poor cross-scene generalization, and generated masked areas often contain blurry artifacts. Predictive filtering is a method for restoring images, which predicts the most effective kernels based on the input scene. Motivated by this approach, we address image completion as a filtering problem. Deep feature-level semantic filtering is introduced to fill in missing information, while preserving local structure and generating visually realistic content. In particular, a Dual-path Cooperative Filtering (DCF) model is proposed, where one path predicts dynamic kernels, and the other path extracts multi-level features by using Fast Fourier Convolution to yield semantically coherent reconstructions. Experiments on three challenging image completion datasets show that our proposed DCF outperforms state-of-art methods.
\end{abstract}
\section{Introduction}
\label{sec:intro}

The objective of image completion (inpainting) is to recover images by reconstructing missing regions. Images with inpainted details must be visually and semantically consistent. Therefore, robust generation is required for inpainting methods. Generative adversarial networks (GANs) \cite{cha2022dam, shamsolmoali2022gen} or auto-encoder networks \cite{peng2021generating, wan2021high, yang2022detail} are generally used in current state-of-the-art models \cite{liu2020rethinking, liu2022reference, suvorov2022resolution} to perform image completion. In these models, the input image is encoded into a latent space by generative network-based inpainting, which is then decoded to generate a new image. The quality of inpainting is entirely dependent on the data and training approach, since the procedure ignores priors (for example smoothness among nearby pixels or features). It should be noted that, unlike the generating task, image inpainting has its own unique challenges. First, image inpainting requires that the completed images be clean, high-quality, and natural. These constraints separate image completion from the  synthesis tasks, which focuses only on naturalness. Second, missing regions may appear in different forms, and the backgrounds could be from various scenes. Given these constraints, it is important for the inpainting method to have a strong capacity to generalize across regions that are missing. Recent generative networks have made substantial progress in image completion, but they still have a long way to go before they can address the aforementioned problems.

\begin{figure}
\centering
\includegraphics[width=0.47\textwidth]{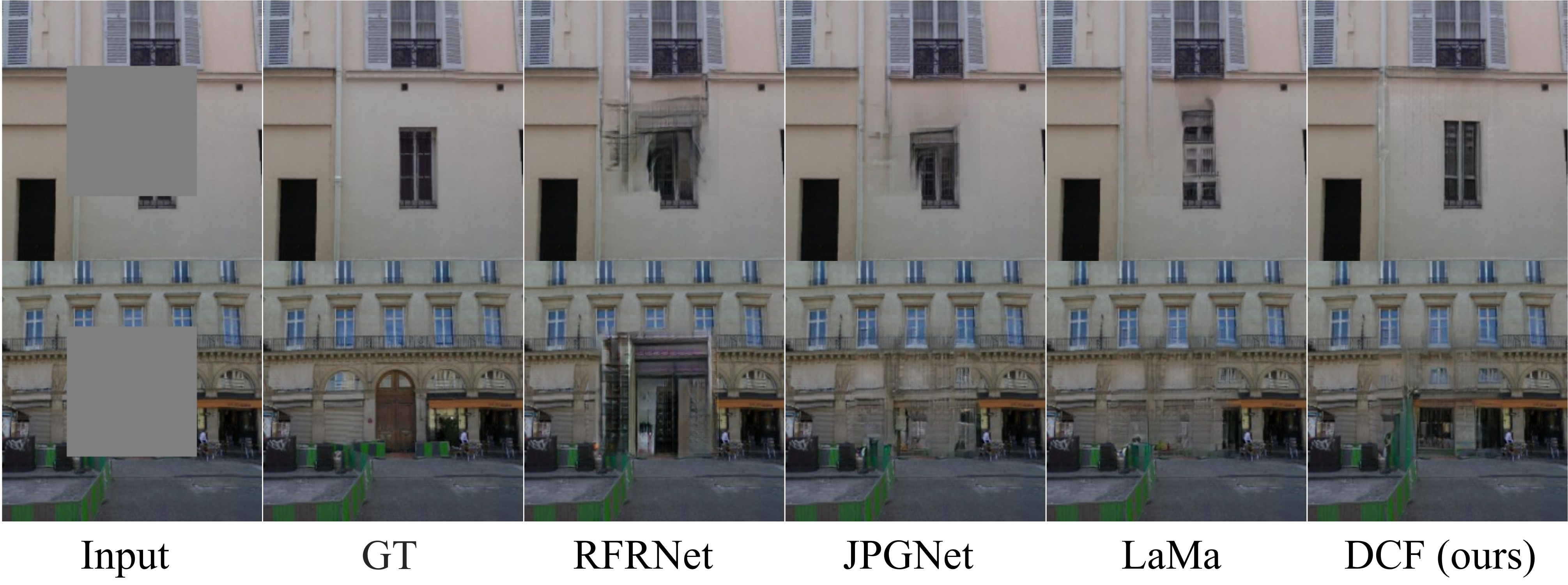}
\caption{Examples of an image completed with our DCF model compared to baseline methods on the Paris dataset. DCF generates high-fidelity and more realistic images.}
\label{paris-comp}
\end{figure}
For instance, RFRNet \cite{li2020recurrent} uses feature reasoning on the auto-encoder architecture for the task of image inpainting. As shown  in Fig. \ref{paris-comp}, RFRNet produces some artifacts in output images.  JPGNet and MISF  \cite{guo2021jpgnet, li2022misf} are proposed to address generative-based inpainting problems \cite{li2020recurrent, lu2022glama, nazeri2019edgeconnect} by reducing artifacts using image-level predictive filtering. Indeed, image-level predictive filtering reconstructs pixels from neighbors, and filtering kernels are computed adaptively based on the inputs. JPGNet is therefore able to retrieve the local structure while eliminating artifacts. As seen in Fig. \ref{paris-comp}, JPGNet's artifacts are more efficiently smoother than RFRNet's. However, many details may be lost, and the actual structures are not reconstructed.
LaMa \cite{suvorov2022resolution} is a recent image inpainting approach that uses Fast Fourier Convolution (FFC) \cite{chi2020fast} inside their ResNet-based LaMa-Fourier model to address the lack of receptive field for producing repeated patterns in the missing areas. Previously, researchers struggled with global self-attention \cite{yu2019free} and its computational complexity, and they were still unable to perform satisfactory recovery for repeated man-made structures as effectively as with LaMa. Nonetheless, as the missing regions get bigger and pass the object boundary, LaMa creates faded structures. In \cite{lu2022glama}, authors adopts LaMa as the base network, and can captures various types of missing information by utilizing additional types of masks. They use more damaged images in the training phase to improve robustness. However, such a training strategy is unproductive.
Transformer-based approaches \cite{wan2021high, zheng2022bridging} recently have attracted considerable interest, despite the fact that the structures can only be estimated within a low-resolution coarse image, and good textures cannot be produced beyond this point. Recent diffusion-based inpainting models \cite{lugmayr2022repaint, rombach2022high} have extended the limitations of generative models by using image information to sample the unmasked areas or use a score-based formulation to generate unconditional inpainted images, however, these approaches are not efficient in real-world applications.

To address this problem, we introduce a new neural network architecture that is motivated by the predictive filtering on adaptability and use large receptive field for producing repeating patterns. In particular, this paper makes two key contributions. First, semantic filtering is introduced to fill the missing image regions by expanding image-level filtering into a feature-level filtering. Second, a Dual-path Cooperative Filtering (DCF) model is introduced that integrates two semantically connected networks -- a kernel prediction network, and a semantic image filtering network to enhance image details. 

The semantic filtering network supplies multi-level features to the kernel prediction network, while the kernel prediction network provides dynamic kernels to the semantic filtering network. In addition, for efficient reuse of high-frequency features, FFC \cite{chi2020fast} residual blocks are utilized in the semantic filtering network to better synthesize the missing regions of an image, leading to improved performance on textures and structures. By linearly integrating neighboring pixels or features, DCF is capable of reconstructing them with a smooth prior across neighbors. Therefore, DCF utilizes both semantic and pixel-level filling for accurate inpainting. Following Fig. \ref{paris-comp}, the propose model produces high-fidelity and realistic images. Furthermore, in comparison with existing methods, our technique involves a dual-path network with a dynamic convolutional operation that modifies the convolution parameters based on different inputs, allowing to have strong generalization.  A comprehensive set of experiments conducted on three challenging benchmark datasets (CelebA-HQ \cite{karras2017progressive}, Places2 \cite{zhou2017places}, and Paris StreetView \cite{doersch2012makes}), shows that our proposed method yields better qualitative and quantitative results than state-of-art methods.
\vspace{-8pt}
\begin{figure}
  \centering
  \includegraphics[width=0.45\textwidth]{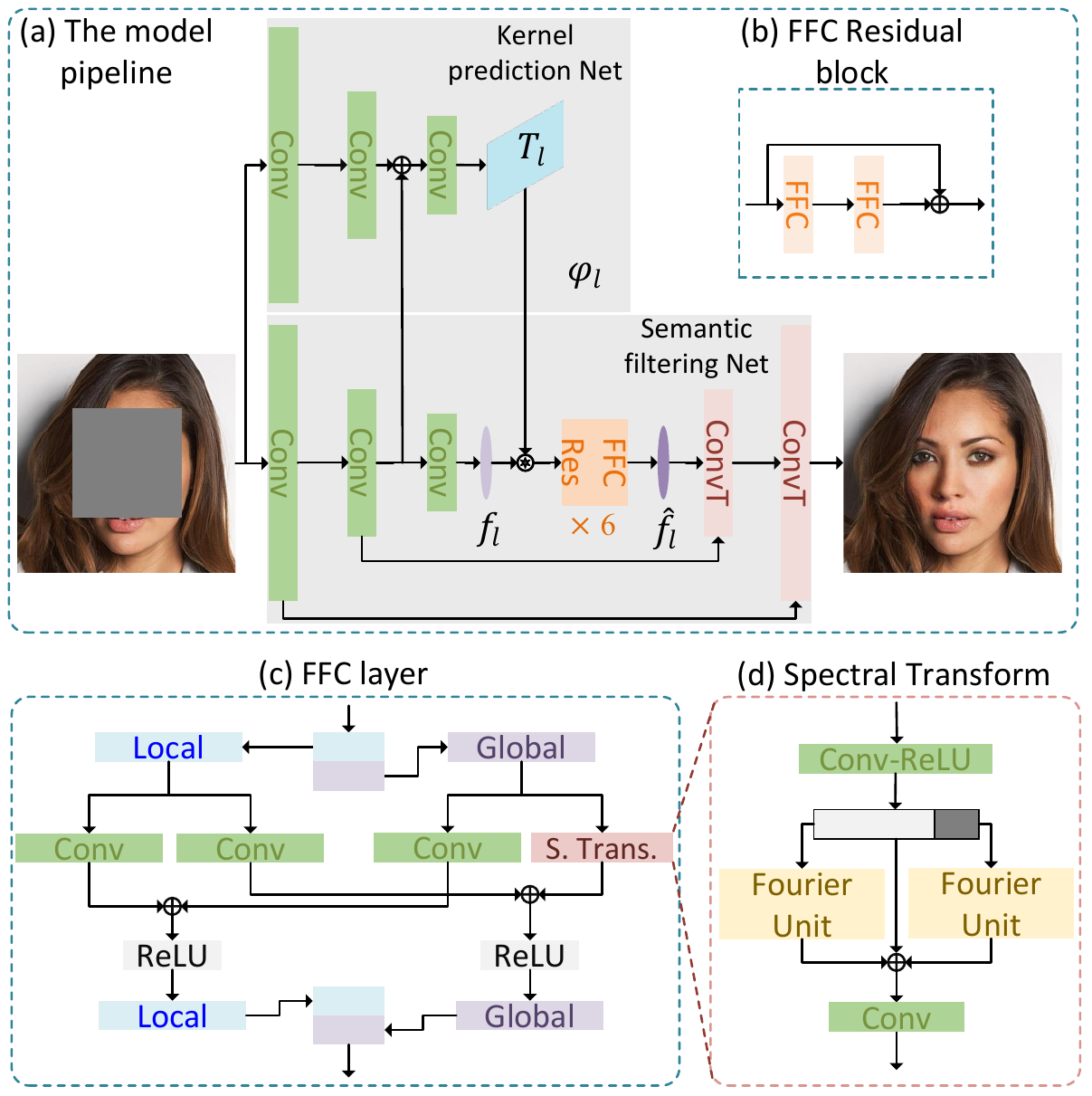}
\caption{Overview of the proposed architecture. (a) Our proposed DCF inpainting network with (b) FFC residual block to have a larger receptive field. (c) and (d) show the architecture of the FFC and Spectral Transform layers, respectively.}
\label{fig:1}
\end{figure}
\section{METHODOLOGY}
\label{sec:format}
Predictive filtering is a popular method for restoring images that is often used for image denoising tasks \cite{mildenhall2018burst}. We define image completion as pixel-wise predictive filtering: 
\begin{equation}
\begin{array}{lr}
	I_c=I_m \ostar T,
    \label{Eq.1}
\end{array}
\end{equation}
\noindent in which $I_c\in \mathbb{R}^{(H\times W\times3)}$ represents a complete image, $I_m\in \mathbb{R}^{(H\times W\times3)}$ denotes the input image with missing regions from the ground truth image $I_{gr}\in \mathbb{R}^{(H\times W\times 3)}$. The tensor $T\in \mathbb{R}^{(H\times W\times N^2)}$ has $HW$ kernels for filtering each pixel and the pixel-wise filtering operation is indicated by the operation $'\ostar'$. Rather than using image-level filtering, we perform the double-path feature-level filtering, to provides more context information. Our idea is that, even if a large portion of the image is destroyed, semantic information can be maintained. To accomplish semantic filtering, we initially use an auto-encoder network in which the encoder extracts features of the damaged image $I_m$, and the decoder maps the extracted features to the complete image $I_c$. Therefore, the encoder can be defined by:
\begin{equation}
\begin{array}{lr}
	f_L=\rho(I_m)=  \rho_L(...\rho_l(...\rho_2(\rho_1(I_m)))),
    \label{Eq.2}
\end{array}
\end{equation}
\noindent in which $\rho(.)$ denotes the encoder while $f_l$ represents the feature taken from the deeper layers ($l^{th}$), $f_l = \rho_{l}(f_{l-1})$. For instance, $f_l$ shows the last layer's result of $\rho(.)$.

In our encoder network, to create remarkable textures and semantic structures within the missing image regions, we adopt Fast Fourier Convolutional Residual Blocks (FFC-Res) \cite{suvorov2022resolution}. The FFC-Res shown in Fig. \ref{fig:1} (b) has two FFC layers. The channel-wise Fast Fourier Transform (FFT) \cite{brigham1967fast} is the core of the FFC layer \cite{chi2020fast} to provide a whole image-wide receptive field. As shown in Fig. \ref{fig:1} (c), the FFC layer divides channels into two branches: a) a local branch, which utilizes standard convolutions to capture spatial information, and b) a global branch, which employs a Spectral Transform module to analyze global structure and capture long-range context. Outputs of the local and global branches are then combined. Two Fourier Units (FU) are used by the Spectral Transform layer (Fig. \ref{fig:1} (d)) in order to capture both global and semi-global features. The FU on the left represents the global context. In contrast, the Local Fourier Unit on the right side of the image takes in one-fourth of the channels and focuses on the semi-global image information. In a FU, the spatial structure is generally decomposed into image frequencies using a Real FFT2D operation, a frequency domain convolution operation, and ultimately recovering the structure via an Inverse FFT2D operation.
\begin{table}
\centering
\small{
\caption{Network architecture of our DCF model. conv(.,.,.) denotes the kernel size, input and output channels.}
\begin{tabular}{l|c|c||l|c}     
\hline
\multicolumn{3}{c||}{Feature extracting network}& \multicolumn{2}{c}{Predicting network}  \\ \cline{1-5}   
Layer&In.&Out./size & In.&Out./size  \\ \hline \hline
conv(7,3,64)  &$I_m$& $f_1$ / 256&  $I_m$& $e_1$ / 256                      \\[-0.5ex]
conv(4,64,128)  &$f_1$& $f_2$ / 128&  $e_1$& $e_2$ / 128    \\ [-0.5ex] 
pooling  &$f_2$& $f^\prime_2$ / 64&  $e_2$& $e^\prime_2$ / 64  \\ [-0.5ex]
conv(4,128,256)  &$f^\prime_2$& $f_3$ / 64&  $[f^\prime_2, e^\prime_2]$& $e_3$ / 64    \\ [-0.5ex]
$f_3\ostar T_3$  &$f_3$& $f^\prime_3$ / 64&  $e_3$& $T_3$ / $64$  \\[-0.5ex]
conv(1,256,256)  &$f^\prime_3$& $f_4$ / 64&  -& -  \\[-0.5ex]
6$\times$FFC  &$f_4$& $f_5$ / 64&  -& -  \\[-0.5ex]
convT(1,256,256)  &$f_5$& $f_6$ / 64&  -& -  \\[-0.5ex]
convT(4,256,128)  &$f_6$& $f_7$ / 64&  -& -  \\[-0.5ex]
convT(4,128,64)  &$f_7$& $f_8$ / 128&  -& -  \\[-0.5ex]
convT(7,64,C)  &$f_8$& $f_9$ / 256&  -& -  \\[-0.4ex]
 \hline
\end{tabular}}
\label{tab:1}
  \end{table}
Therefore, based on the encoder the network of our decoder is defined as:
\begin{equation}
\begin{array}{lr}
	I_c=\rho^{-1}(f_L),
    \label{Eq.3}
\end{array}
\end{equation}
\noindent in which $\rho^{-1}(.)$ denotes the decoder. Then, similar to image-level filtering, we perform semantic filtering on extracted features according to:
\begin{equation}
	\hat f_l[r] =\sum_{s\in \mathcal{N}_\kappa} {T^l_\kappa[s-r]f_l[s]},
    \label{Eq.4}
\end{equation}
\noindent in which $r$ and $s$ denote the image pixels' coordinates, whereas the $\mathcal{N}_\kappa$ consist of $N^2$ closest pixels.
$T^l_\kappa$ signifies the kernel for filtering the $\kappa^{th}$ component of $T_l$ through its neighbors $\mathcal{N}_{\kappa}$.
To incorporate every element-wise kernel, we use the matrix $T_l$ as $T^l_\kappa$. Following this, Eq. (\ref{Eq.2}) is modified by substituting $f_l$ with $\hat f_l$. In addition, we use a predictive network to predict the kernels' behaviour in order to facilitate their adaptation for two different scenes.
\begin{equation}
	T_l =\varphi_l (I_m),
    \label{Eq.5}
\end{equation}
\noindent in which $\varphi_l (.)$ denotes the predictive network to generate $T_l$.
In Fig. \ref{fig:1}(a) and Table \ref{tab:1}, we illustrate our image completion network which consist of $\rho(.), \rho^{(-1)},$ and $\varphi_l(.)$. The proposed network is trained using the $L_1$ loss, perceptual loss, adversarial loss, and style loss, similar to predictive filtering.
\begin{figure}[t]
  \centering
  \includegraphics[width=0.45\textwidth]{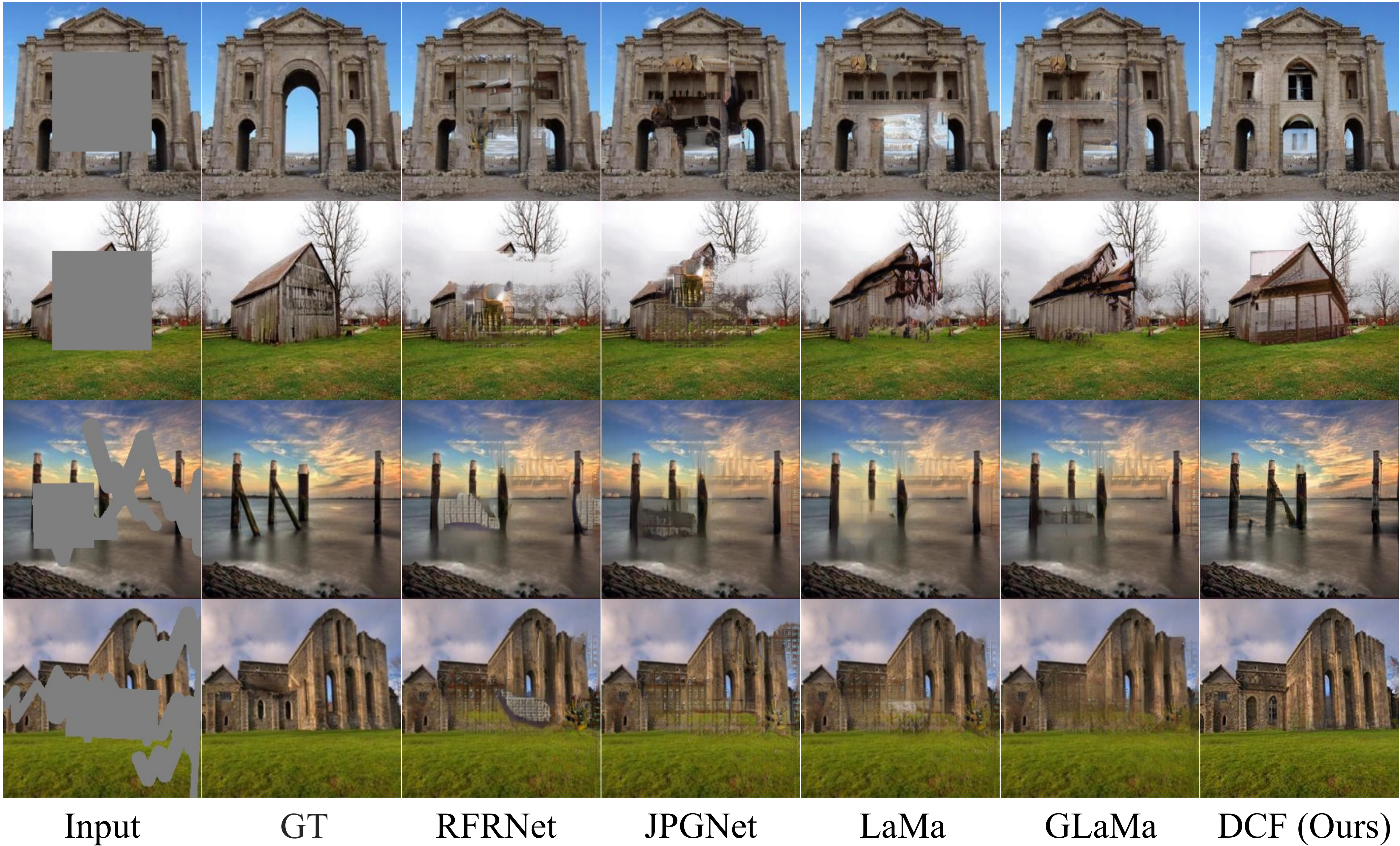}
\vspace{-6pt}
\caption{Qualitative comparison on the Places2 dataset. Our model outperforms state-of-art methods in terms of both structure and texture preservation.}
\label{placeR-comp}
\end{figure}

\begin{figure}[t]
  \centering
  \includegraphics[width=0.45\textwidth]{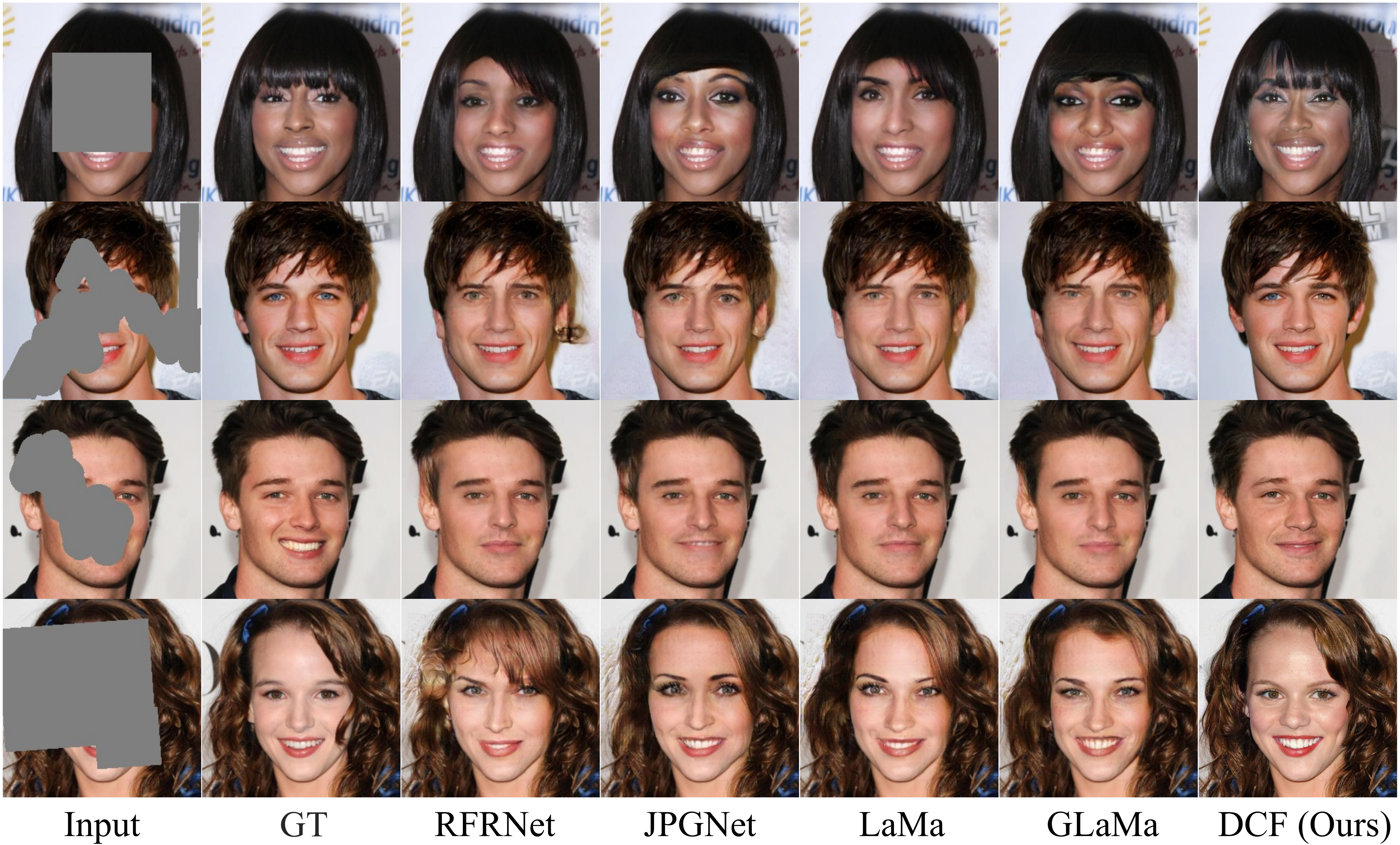}
\vspace{-6pt}
\caption{Qualitative comparison on CelebA data. Facial images produced by DCF are more realistic, and have more characteristic facial features compared to state-of-art methods.}
\label{celeb-comp}
\end{figure}

\newcommand{\mycc}{\cellcolor{LightCyan}}
\begin{table*}
\renewcommand{\arraystretch}{1.1} 
\centering
\small{
\caption{Ablation study and quantitative comparison of our proposed and state-of-art methods on center and free form masked images from the CelebA-HQ, Places2, and Paris StreetView datasets.} \label{tab:2}
\begin{tabular}{c|l|cc|cc|cc}     \hline\hline
\rule{0pt}{1\normalbaselineskip}  & \multirow{ 2}{*}{Method} & \multicolumn{2}{c|}{CelebA-HQ}& \multicolumn{2}{c|}{Places2}& \multicolumn{2}{c}{Paris StreetView}  \\ [-0.3ex]  \cline{3-8}   
             & \rule{0pt}{1\normalbaselineskip}& Irregular & Center& Irregular & Center& Irregular & Center  \\ [-0.3ex]   \hline

\rule{0pt}{1\normalbaselineskip} 
\multirow{ 4}{*}{PSNR$\uparrow$}& RFRNet \cite{li2020recurrent} &   26.63& 21.32& 22.58 & 18.27& 23.81& 19.26  \\  [-0.8ex]
&JPGNet \cite{guo2021jpgnet} &   25.54& 22.71& 23.93 & 19.22& 24.79& 20.63   \\ [-0.8ex]
&TFill \cite{zheng2022bridging} &   26.84& 23.65& 24.32 & 20.49&25.46  &21.85  \\  [-0.8ex]
&LaMa \cite{suvorov2022resolution} & 27.31& 24.18& \bf{25.27} & 21.67& 25.84& 22.59 \\ [-0.8ex]
&GLaMa \cite{lu2022glama}  &   {28.17}& 25.13& 25.08 & 21.83&26.23  & 22.87 \\ [-0.8ex]
&\mycc DCF (ours) &\mycc    \bf{28.34}&\mycc \bf{25.62}&\mycc  25.19 &\mycc \bf{22.30}&\mycc \bf{26.57}&\mycc  \bf{23.41} \\   \hline
\multirow{ 4}{*}{SSIM$\uparrow$}&RFRNet \cite{li2020recurrent} &   0.934& 0.912& 0.819 & 0.801& 0.862& 0.849  \\  [-0.8ex]
&JPGNet \cite{guo2021jpgnet} &   0.927& 0.904& 0.825 & 0.812& 0.873& 0.857   \\ [-0.8ex]
&TFill \cite{zheng2022bridging} &   0.933& 0.907& 0.826 & 0.814&0.870  &0.857  \\  [-0.8ex]
&LaMa \cite{suvorov2022resolution} &  0.939& 0.911& 0.829 & 0.816& 0.871& 0.856    \\                                    [-0.8ex]
&GLaMa \cite{lu2022glama}  &  0.941& 0.925& \bf{0.833} & 0.817& 0.872  & 0.858 \\ [-0.8ex]
&\mycc DCF (ours) &\mycc  \bf{0.943}&\mycc \bf{0.928}&\mycc  0.832 &\mycc \bf{0.819}&\mycc \bf{0.876}&\mycc  \bf{0.861} \\   \hline
\multirow{ 4}{*}{FID$\downarrow$}&RFRNet \cite{li2020recurrent} &  17.07& 17.83  & 15.56& 16.47& 40.23 & 41.08  \\  [-0.8ex] 
&JPGNet \cite{guo2021jpgnet} &    13.92& 15.71& 15.14 & 16.23& 37.61& 39.24   \\ [-0.8ex]
& TFill \cite{zheng2022bridging} &  13.18& 13.87  & 15.48& 16.24&33.29  &34.41  \\  [-0.8ex] 
&LaMa \cite{suvorov2022resolution} &   11.28& 12.95& 14.73 & 15.46& 32.30& 33.26    \\                                                      [-0.8ex]
&GLaMa \cite{lu2022glama} &    {11.21}& 12.91& 14.70 & 15.35&32.12   & 33.07  \\ [-0.8ex]
\rule{0pt}{0.8\normalbaselineskip} 
&\mycc DCF w.o. Sem-Fil & \mycc  14.34& \mycc 15.24&\mycc 17.56 &\mycc 18.11&\mycc 42.57 &\mycc 44.38   \\ [-0.8ex]
&\mycc DCF w.o. FFC & \mycc  13.52& \mycc 14.26& \mycc15.83 & \mycc16.98&\mycc 40.54  &\mycc 41.62   \\ [-0.8ex] 
&\mycc DCF (ours) &\mycc \bf{11.13}& \mycc \bf{12.63}& \mycc \bf{14.52} & \mycc\bf{15.09}&\mycc \bf{31.96} &\mycc \bf{32.85} \\  [-0.5ex] \hline
\end{tabular}}
\end{table*}
\section{EXPERIMENTS}
\label{sec:pagestyle}
In this section, the performance of our DCF model is compared to state-of-the-art methods for image completion task. Experiments are carried out on three datasets, CelebA-HQ \cite{karras2017progressive}, Places2 \cite{zhou2017places}, and Paris StreetView \cite{doersch2012makes} at $256\times 256$ resolution images. With all datasets, we use the standard training and testing splits. In both training and testing we use the diverse irregular mask (20\%-40\% of images occupied by holes) given by PConv \cite{liu2018image} and regular center mask datasets. The code is provided at \href{https://github.com/pshams55/DCF}{\it {DCF}}.\\
\vspace{-9pt}

\noindent {\bf{Performance Measures:}} The structural similarity index (SSIM), peak signal-to-noise ratio (PSNR), and Frechet inception distance (FID) are used as the evaluation metrics.

\subsection{Implementation Details}

Our proposed model's framework is shown in Table \ref{tab:1}.\\
\vspace{-8pt}

\noindent {\bf{Loss functions.}} We follow \cite{nazeri2019edgeconnect} and train the networks using four loss functions, including $L_1$ loss ($\ell_{1}$), adversarial loss ($\ell_{A}$), style loss ($\ell_{S}$), and perceptual loss ($\ell_{P}$), to obtain images with excellent fidelity in terms of quality as well as semantic levels. Therefore, we can write the reconstruction loss ($\ell_{R}$) as:
\begin{equation}
\ell_{R}=\lambda_{1}\ell_{1}+\lambda_{a}\ell_{A}+\lambda_{p}\ell_{P}+\lambda_{s}\ell_{S}.
\label{eq:2-9}
\end{equation}
in which $\lambda_{1}=1$, $\lambda_{a}=\lambda_{p}=0.1$, and $\lambda_{s}=250$. More details on the loss functions can be found in \cite{nazeri2019edgeconnect}.\\
\vspace{-8pt}

\noindent {\bf{Training setting.}} We use Adam as the optimizer with the learning rate of $1e-4$ and the standard values for its hyper-parameters. The network is trained for 500k iterations and the batch size is 8. The experiments are conducted on the same machine with two RTX-3090 GPUs.
\subsection{Comparisons to the Baselines}

\noindent {\bf{Qualitative Results.}} The proposed DCF model is compared to relevant baselines such as RFRNet \cite{li2020recurrent}, JPGNet \cite{guo2021jpgnet}, and LaMa \cite{suvorov2022resolution}. Fig. \ref{placeR-comp} and Fig. \ref{celeb-comp}  show the results for the Places2 and CelebA-HQ datasets respectively.  In comparison to JPGNet, our model preserves substantially better recurrent textures, as shown  in Fig. \ref{placeR-comp}. Since JPGNet lacks attention-related modules, high-frequency features cannot be successfully utilized due to the limited receptive field. Using FFC modules, our model expanded the receptive field and successfully project source textures on newly generated structures. Furthermore, our model generates superior object boundary and structural data compared to LaMa. Large missing regions over larger pixel ranges limit LaMa from hallucinating adequate structural information. However, ours uses the advantages of the coarse-to-fine generator to generate a more precise object with better boundary. Fig. \ref{celeb-comp} shows more qualitative evidence. While testing on facial images, RFRNet and LaMa produce faded forehead hairs and these models are not robust enough. The results of our model, nevertheless, have more realistic textures and plausible structures, such as forehead form and fine-grained hair.\\
\vspace{-8pt}

\noindent {\bf{Quantitative Results.}} On three datasets, we compare our proposed model with other inpainting models. The results shown in Table \ref{tab:2} lead to the following conclusions: 1) Compared to other approaches, our method outperforms them in terms of PSNR, SSIM, and FID scores for the most of datasets and mask types. Specifically, we achieve 9\% higher PNSR on the Places2 dataset's irregular masks than RFRNet. It indicates that our model has advantages over existing methods. 2) We observe similar results while analyzing the FID. On the CelebA-HQ dataset, our method achieves 2.5\% relative lower FID than LaMa under the center mask. This result indicates our method's remarkable success in perceptual restoration. 3) The consistent advantages over several datasets and mask types illustrate that our model is highly generalizable.
\vspace{-8pt}
\section{Conclusion}
\label{sec:5}
Dual-path cooperative filtering (DCF) was proposed in this paper for high-fidelity image inpainting. For predictive filtering at the image and deep feature levels, a predictive network is proposed. In particular, image-level filtering is used for details recovery, whereas deep feature-level filtering is used for semantic information completion.
Moreover, in the image-level filtering the FFC residual blocks is adopted to recover semantic information and resulting in high-fidelity outputs. The experimental results demonstrate our model outperforms the state-of-art inpainting approaches. 
\vspace{-8pt}
\section*{\small{Acknowledgments}}
\vspace{-6pt}
\noindent This research was supported in part by NSFC China. The corresponding author is Masoumeh Zareapoor.

\bibliographystyle{plain}
\bibliography{strings,refs}

\end{document}